\documentclass[runningheads]{llncs}
\usepackage{graphicx}
\begin{document}
\title{A Comparative Study of Machine Learning and Deep Learning Techniques for Prediction of Co2 Emission in Cars}
\titlerunning{Prediction of CO2 emissions in cars}
%
\author{Samveg Shah\inst{0000-0003-3183-5046} \and
Shubham Thakar\inst{0000-0003-3101-9735} \and
Kashish Jain\inst{0000-0002-1115-2507}
\and
Bhavya Shah\inst{0000-0002-4280-9786}
\and
Sudhir Dhage\inst{0000-0001-8100-8941}}

\authorrunning{Samveg, Shubham et al.}

\institute{Sardar Patel Institute of Technology, Mumbai, India
\\
\email{\{samveg.shah, shubham.thakar, kashish.jain, bhavya.shah, sudhir\_dhage\}@spit.ac.in}}
\maketitle              
\begin{abstract}
The most recent concern of all people on the Earth is the increase in the concentration of greenhouse gas in the atmosphere. The concentration of these gases has risen rapidly over the last century and if the trend continues it can cause many adverse climatic changes. There have been ways implemented to curb this by the government by limiting processes that emit a higher amount of CO2, one such greenhouse gas. However, there is mounting evidence that the CO2 numbers supplied by the government do not accurately reflect the performance of automobiles on the road. Our proposal of using artificial intelligence techniques to improve a previously rudimentary process takes a radical tack, but it fits the bill given the situation. To determine which algorithms and models produce the greatest outcomes, we compared them all and explored a novel method of ensembling them. Further, this can be used to foretell the rise in global temperature and to ground crucial policy decisions like the adoption of electric vehicles. To estimate emissions from vehicles, we used machine learning, deep learning, and ensemble learning on a massive dataset.

\end{abstract}
\section{Introduction}
The rise of personal transport in recent years has been accompanied by a sharp increase in human CO2 emissions across the globe. About a quarter of the EU's total CO2 emissions in 2019 were attributable to transportation, the majority (71.7\%) of which originated from vehicle travel. Greenhouse gas emissions have increased only in the transportation sector over the past three decades, increasing by 33.5 percent from 1990 to 2019.

Significantly reducing CO2 emissions from transport will not be easy as the rate of emission reductions has slowed. Although the goal is to reduce emissions from transportation by 80\% by 2050, current projections put that number much lower, at only 22\%. Predictions for the year 2050 indicate a much lower reduction in transportation-related emissions than what is currently desired. There is a large disparity between the CO2 emissions of passenger transport modalities. Sixty-one percent of all carbon dioxide emissions from automobiles in the European Union are caused by personal vehicles. \cite{ref_link1}

According to official laboratory measured monitoring statistics, the average fuel consumption and CO2 emissions of the European passenger car fleet have been steadily decreasing. There is mounting evidence that the CO2 numbers supplied by the government do not accurately reflect the performance of automobiles on the road. The gap between published figures and real-world estimations was believed to be 30–40 percent and has been widening over time. Between 10 and 20 percent of the discrepancy between the stated values and reality can be attributed to the margins of the current certification method. For average fleet emissions in 2015, the latter was calculated to be around 40\%, or 47.5 gCO2/km. However, it may be as high as 60\% or as low as 19\% depending on the day's traffic. \cite{ref_link2}

Even while work is being done to reduce these gaps, there is still a severe lack of alternate approaches to estimating emissions. This proposal takes a radical tack, but it's one that fits the bill given the gravity of the situation. In order to demonstrate how these cutting-edge methods might offer a fresh perspective on the problem at hand, we have used machine learning and deep learning to analyze a sizable collection of vehicle data.

The paper begins with a review of the relevant literature survey in section 2, then moves on to describe the dataset used and the preprocessing done in section 3. Then it moves on to describe the methodologies used in section 4. Then it presents the results in section 5 and ends with the conclusion in section 6.

\section{Literature Survey}
In paper \cite{ref_article1} the authors have created a regression analysis model to estimate real-world CO2 emissions of light-duty diesel vehicles. Vehicle speed, engine speed, engine power, and acceleration were some variables that were used for regression analysis. Further CO2 emission estimators were derived from the regression equation and were compared with real CO2 emissions.

The authors of \cite{ref_article2} have made an effort to use the ensemble technique to construct a vehicle trip CO2 model by using a significant amount of lab-based data. They aim to help planners develop better mobility maps that can help reduce the overall carbon footprint.

The aim of the authors in \cite{ref_article3}, is to forecast vehicle CO2 emissions per kilometer and identify an environmentally friendly route that has the lowest CO2 emissions while still meeting trip time constraints. The theory of vehicle dynamics is the foundation from which the vehicle CO2 emission model was constructed.

The authors of paper \cite{ref_article4} used data mining techniques to examine the remote sensing data and estimate the type of fuel and the registration time. Using decision trees, random forests, AdaBoost, XGBoost, and their fusion models, the study's objective was to accurately predict these two important pieces of information and then apply those predictions to a critical application, namely the evaluation of automobile emissions.

The variety of different cars and the ongoing advancement of vehicle exhaust purification technology make it difficult to estimate vehicular transient emissions accurately, as stressed by paper \cite{ref_article5}. They have built CO2 and NOx transient emission models using a Super-learner model to precisely characterize the transient emissions of motor vehicles.

\cite{ref_article6} considers various tabular datasets and compares the use of deep learning models and XGBoost on them. They concluded that XGBoost outperforms deep models on all datasets and also requires much less tuning. However, an ensemble of XGBoost and deep models performed better than XGBoost alone. This motivated us to utilize an ensemble of deep models and XGBoost to tackle the problem of predicting CO2 emissions in cars.

By taking into account both linear and non-linear methods for time-series forecasting, as well as ensemble and hybrid models by combining components from various alternatives, the author of the paper [22] surveys the most recent advancements in supervised machine learning (ML) and high dimensional models. They also use time series forecasting in the financial and economic sectors. In [23], the author examines a variety of univariate models, including AutoRegressive (AR) models, to determine how well they can predict the future.

The study [24] applied and compared eight different machine learning models to the well-known M3 time series competition data. The multilayer perceptron and the regression based on the gaussian process fared the best. The authors of [25] employed time series data of CO2 69 emissions in India from 1960 to 2017 and the Box- Jenkins ARIMA technique. Based on the forecast, they recommended five policy recommendations to enhance the environmental circumstances.

In order to anticipate the Air quality index, [26] conducted a prediction-based study in which they gathered data from Delhi and the National Capital Region (NCR) of India. They used machine learning models for the same purpose and evaluated the results using the MAE, MSE, RMSE, and MAPE metrics. 

\section{About The Dataset}
The dataset we used was found on the European union website, which had over 90,00,000 rows and 33 different columns. Due to the humongous size of the dataset, we decided to do some preprocessing first. We eliminated rows with null values so that it wouldn't have any adverse effect on our model training. Then we analyzed the correlation matrix and decided on columns that were important and not important to the dataset. The dataset we eventually ended up getting had close to 10,00,000 rows and 18 columns, which have been used to train the model.

\section{Methodology}
\begin{enumerate}
  \item \textit{Preprocessing} \\
    The min-max normalization which is a very common technique for normalization was used on the dataset (excluding the target variable). For each feature, the minimum value is converted to a 0, the maximum value is converted to a 1, and all other values are converted to a decimal between 0 and 1. This normalization technique helps avoid the problem of exploding gradients in the neural network. \par
    We have used the One hot encoding technique to encode nominal features of the dataset since there is no numerical relationship between the different categories. In the case of One hot encoding, each integer value is represented by a binary vector. The features which have an inherent ranking within the data points are called ordinal features and ordinal encoding is used for such categorical features. After splitting the data the dataset has 1,00,000 rows, while the training dataset has 9,00,000 rows.
    \begin{figure}[h]
    
    \centering
    \includegraphics[width=0.5\paperwidth]{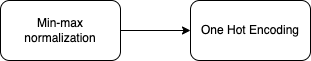}
    \caption{Steps for preprocesing}
    \end{figure}
  \item \textit{Models}
    \begin{enumerate}
        \item \textit{Random Forest Regressor}
                    \begin{figure}[h]
                    
                    \centering
                    \includegraphics[width=0.5\paperwidth]{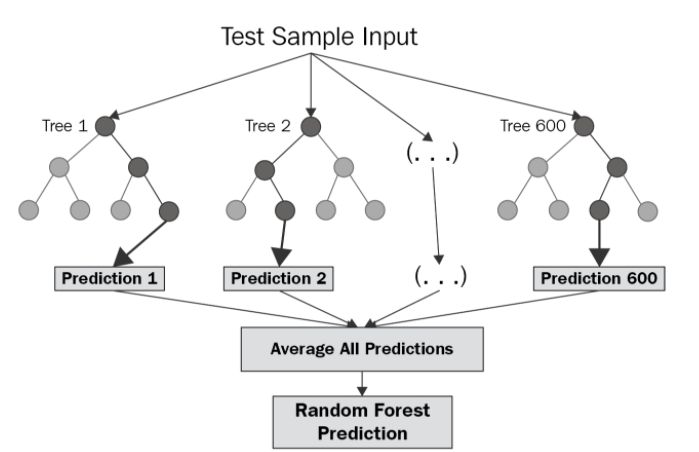}
                    \caption{Example of a random forest regressor}
                    \end{figure}
            \begin{enumerate}
                \item Random Forest is an ensemble technique that can perform both regression and classification tasks by combining multiple decision trees and a technique known as Bootstrap and Aggregation, or bagging. The basic idea is to use multiple decision trees to determine the final output rather than relying on individual decision trees.\cite{ref_link4}
                \item The random forest regressor parameters that produced the best results had n\_estimators (the number of decision trees in the model) set to 250, criterion (this variable allows you to choose the criterion (loss function) used to determine model outcomes) set to "absolute" to represent the mean absolute error, and max depth (this variable sets the maximum possible depth of each tree) set to 9.
                \item The MAE ( mean absolute error )  given by this model was 0.65

            \end{enumerate}
        \item \textit{XGBboost}
                    \begin{figure}
                    
                    \centering
                    \includegraphics[width=0.5\paperwidth]{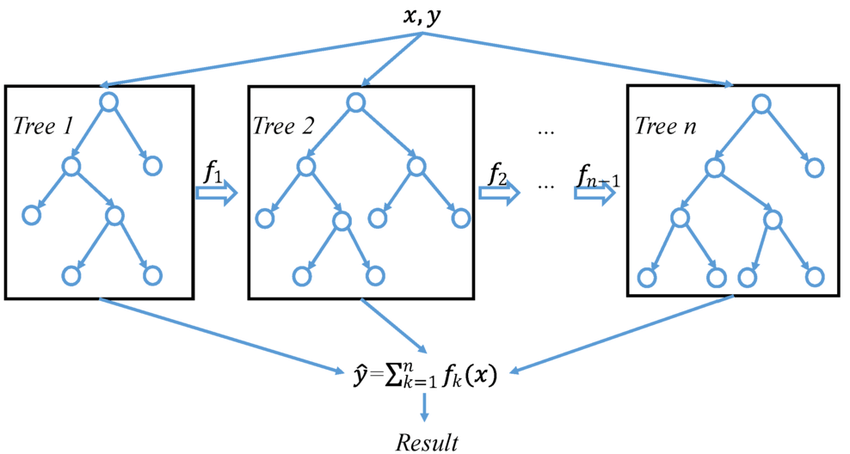}
                    \caption{General architecture of XGBboost}
                    \end{figure}
           \begin{enumerate}
                \item XGBoost is a gradient boosting ensemble Machine Learning algorithm based on decision trees. In prediction issues involving unstructured data, artificial neural networks outperform all existing algorithms or frameworks (images, text, etc.). However, for small to medium amounts of structured/tabular data, decision tree-based algorithms are currently regarded as best-in-class. \cite{ref_link5}
                \item The XGBboost regressor parameters which gave the most optimized results had the n\_estimators (This is the total number of estimators added during model training.) set to 1000, learning\_rate ( The rate at which the XGBoost model learns during the training phase)  set to 0.05,  and objective='reg: linear' ( It specifies the type of algorithm used to build the model ) set to 'reg: linear'. These parameters which gave the most optimized results were based on a randomized search.
                \item The MAE ( mean absolute error )  given by this model was 2.73

            \end{enumerate}
        \item \textit{Neural Network}
                    \begin{figure}[h]
                    
                    \centering
                    \includegraphics[width=0.5\paperwidth]{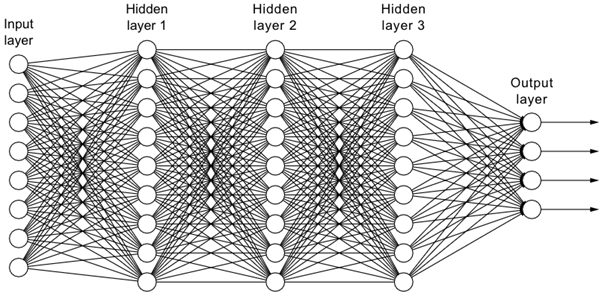}
                    \caption{Sample architecture of neural network}
                    \end{figure}
           \begin{enumerate}
                \item Neural networks, which use neurons to transmit informational bits, closely resemble the way the human brain functions. Artificial neurons used in neural networks combine multiple inputs to generate a single output. There are three layers: Input, Hidden, and Output \cite{ref_link7}. The hidden layers perform transformations on the input and are specialized to produce a definite output.
                \item We have used the rectified linear activation function with all the layers. Some advantages of using ReLu include ease of computation, and linear behavior [27].  The sequential model employed in our neural network architecture has the following layers:
                           \begin{enumerate}
                           \item Dense layer having 128 neurons and ReLu activation function 
                           \item Dense layer having 512 neurons and ReLu activation function
                           \item Dense layer having 512 neurons and ReLu activation function
                           \item Dense layer having 512 neurons and ReLu activation function
                           \item  Dense layer having 256 neurons and ReLu activation function
                           \item Dense layer having 256 neurons and ReLu activation function
                           \item Dense layer having 1 neuron and ReLu activation function
                           \end{enumerate}
 
The last layer represents the output which is the amount of co2 emission

                \item The Adam optimizer was used to create the model. The results of the Adam optimizer are generally better than every other optimization algorithm, have faster computation time, and require fewer parameters for tuning so we decided to use that. Adaptive Moment Estimation is a technique for optimizing gradient descent algorithms. The method is extremely efficient when dealing with large problems involving a large number of data or parameters. It uses less memory and is more efficient. It appears to be a hybrid of the 'gradient descent with momentum' and the 'RMSP' algorithms. The mean absolute error metric was used.
                \item The model was trained for 500 epochs with a batch size of 128 and a validation split of 20
                \item Total trainable parameters were 991873
                \item The MAE ( mean absolute error )  given by this model was 0.62
            \end{enumerate}
        \item \textit{Ensemble Learning}
                    \begin{figure}[h]
                    \centering
                    \includegraphics[width=0.7\paperwidth]{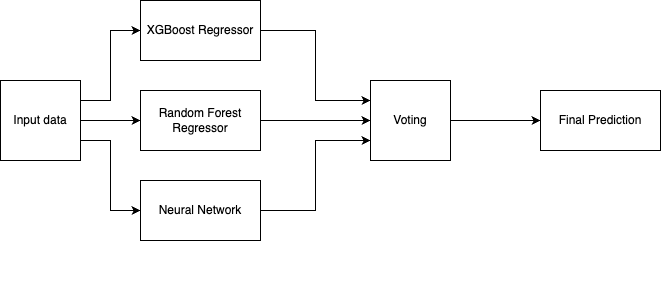}
                    \caption{Sample architecture of voting regressor}
                    \end{figure}
         \begin{enumerate}
                \item Ensemble learning is a method of reaching a prediction consensus by combining the key features of two or more models. Because ensembling reduces the variance in prediction errors, the final ensemble learning framework is more robust than the individual models that comprise the ensemble. There are two primary, linked reasons to choose an ensemble over a single model: Performance: Compared to a single contributing model, an ensemble can anticipate events more accurately and perform better overall. Robustness: An ensemble narrows the prediction and model performance distribution. An ensemble framework is successful when the contributing models are statistically diverse. \cite{ref_link6}
                \item The following are some of the benefits of ensemble modeling for neural networks:
                \begin{enumerate}
                    \item Enhancement of Performance - An ensemble model can yield outcomes with more generalizability and, in most situations, even greater accuracy by combining the prediction capacity of its ensemble members.
                    \item Enhancement of Reliability - Single, base models frequently suffer from variance, bias, and noise issues. The ensemble model improves reliability and robustness by reducing bias and variance and weighting a greater number of features. Robustness represents low variance, indicating that the model is unaffected by minor inconsistencies in the training data. Even if the model's accuracy does not improve, the ensemble reduces the variance or spread of prediction.
                \end{enumerate}

                \item A voting regressor, an ensemble meta-estimator that fits multiple base regressors, each on the entire dataset, is what we utilized. The ultimate prediction is created by averaging the individual predictions.
                \item We experimented with 3 ensembles :
                    \begin{enumerate}
                        \item Random Forest and Neural Networks: This ensemble had an accuracy of 0.58
                        \item Neural Network and XGBoost: This ensemble had an accuracy of 2.2
                        \item Random Forest, XGBoost, Neural Networks: This ensemble had an accuracy of 1.74
                    \end{enumerate}
                \item It can be observed that the ensemble between Random Forest and Neural Network had the highest accuracy among all the ensembles.

            \end{enumerate}
    \end{enumerate}
\end{enumerate}

\section{Results}
As described in the previous sections, we have tested our datasets on various ML algorithms and deep learning techniques. We obtained results in the form of the Mean Absolute Error which is MAE. The formula for the same is

\begin{equation}
\frac{\sum_{i=1}^{n} |y_i-x_i|)}{n} = MAE
\end{equation}

\noindent Where, y\textsubscript{i} is the prediction, x\textsubscript{i} is the true value and n specifies the total number of data points.

\begin{table}
\centering
\caption{Mean Absolute Error Of Different Methods}\label{tab1}
\begin{tabular}{|l|l|l|}
\hline
Method                     & Mean Absolute Error          \\ \hline
Random Forest                                    & 0.65      \\
XGBoost                                            & 2.73    \\
Neural Network                                     & 0.62         \\
Neural Network And XGBoost Ensemble                & 2.2         \\
Neural Network And Random Forest Ensemble          & 0.58      \\
Neural Network,XGBoost,Random Forest Ensemble      & 1.74         \\ \hline
\end{tabular}
\end{table}

\section{Conclusion}
Predicting the emissions of vehicles is a difficult task since it varies greatly with the size, engine capacity, fuel type, etc. However, it is a significant problem to solve as the results can help in predicting overall vehicular emissions and their impact. This can further be used to predict the rise in global temperature and also anchor important policy decisions such as shifting to electric vehicles. We applied machine learning, deep learning, and ensemble learning to a high-volume dataset to solve the problem of predicting vehicular emissions. We got the best results for our neural network and random forest ensemble model for which the MAE stands at 0.58. This proposed deep learning and machine learning ensemble technique can be further tried on other regression/classification problems.


%
%
%
%

\end{document}